\documentclass[10pt,twocolumn,letterpaper]{article}
 
\usepackage[pagenumbers]{cvpr} 

\usepackage{graphicx}
\usepackage{amsmath}
\usepackage{amssymb}
\usepackage{booktabs}
\usepackage{float}

\usepackage[pagebackref,breaklinks,colorlinks]{hyperref}
\usepackage{multirow}
\usepackage{bbding}

\newcommand{\tool}{\textsc{DBMark}\xspace}
\usepackage[capitalize]{cleveref}
\crefname{section}{Sec.}{Secs.}
\Crefname{section}{Section}{Sections}
\Crefname{table}{Table}{Tables}
\crefname{table}{Tab.}{Tabs.}

\def\cvprPaperID{8922} 
\def\confName{CVPR}
\def\confYear{2023}

\begin{document}

\title{Deep Boosting Robustness of DNN-based Image Watermarking via \tool}

\author{
 \small Guanhui Ye$^*$ ~~Jiashi Gao$^*$ ~~Wei Xie$^\circ$ ~~Bo Yin$^\otimes$~~Xuetao Wei$^*$ 
\\
	{\small{$^*$}Southern University of Science and Technology} ~~{\small{$^\circ$}{Hunan University}} ~~{\small{$^\otimes$}{Changsha University of Science and Technology}}\\
}

\maketitle

\begin{abstract}

Image watermarking is a technique for hiding information into images that can withstand distortions while requiring the encoded image to be perceptually identical to the original image. Recent work based on deep neural networks (DNN) has achieved impressive progression in digital watermarking.
Higher robustness under various distortions is the eternal pursuit of digital image watermarking approaches.
In this paper, we propose \tool, a novel end-to-end digital image watermarking framework to deep boost the robustness of DNN-based image watermarking. The key novelty is the synergy of invertible neural networks (INN) and effective watermark features generation. The framework generates watermark features with redundancy and error correction ability through the effective neural network based message processor, synergized with the powerful information embedding and extraction abilities of INN to achieve higher robustness and invisibility. The powerful learning ability of neural networks enables the message processor to adapt to various distortions. In addition, we propose to embed the watermark information in the discrete wavelet transform (DWT) domain and design low-low ($LL$) sub-band loss to enhance invisibility. Extensive experiment results demonstrate the superiority of the proposed framework compared with the state-of-the-art ones under various distortions such as dropout, cropout, crop, Gaussian filter, and JPEG compression.

\end{abstract}


\begin{figure*}[ht]
 \centering
  \includegraphics[width=\linewidth]{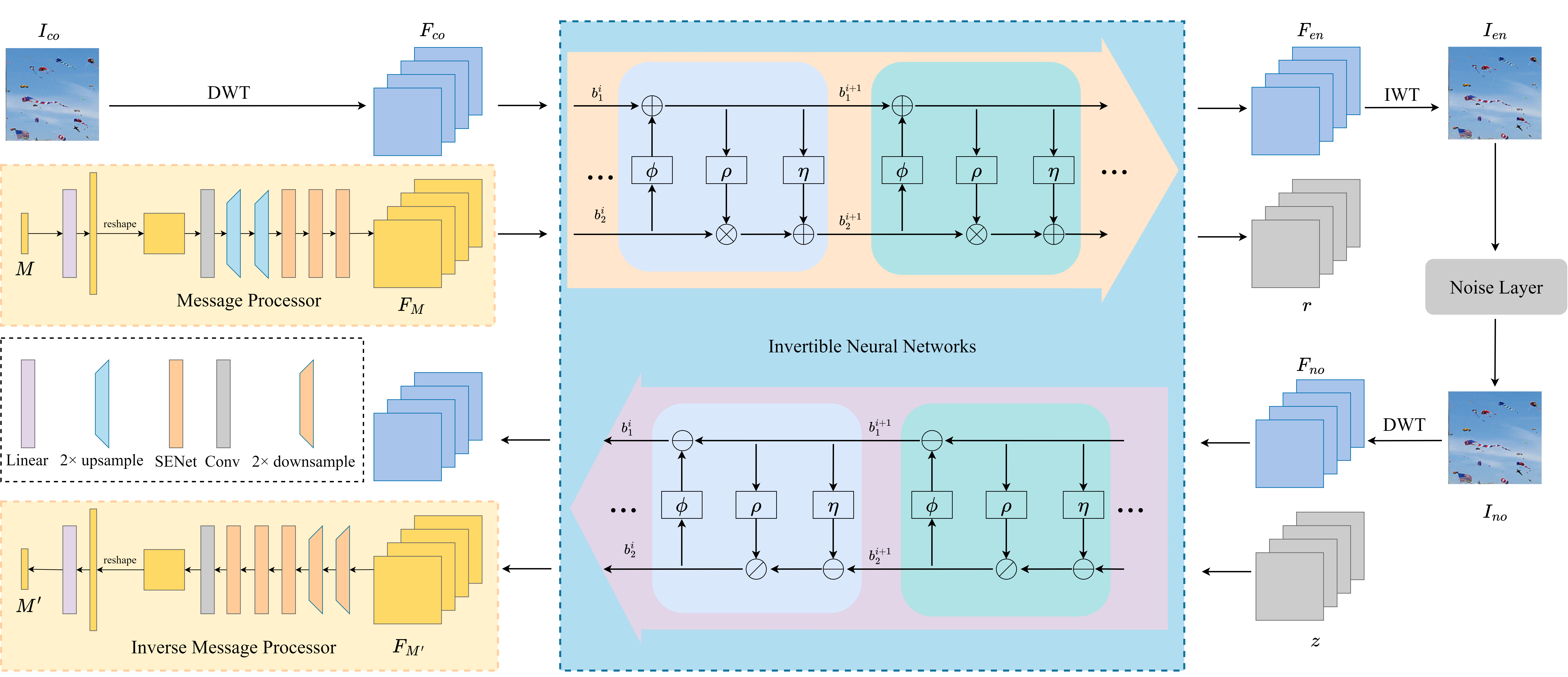}
  \caption{The architecture of \tool. In the watermark encoding process, a secret message $M$ is encoded into a cover image $I_{co}$ to generate an encoded image $I_{en}$. In watermark decoding process, the encoded image $I_{en}$ passing through the noise layer is fed in a reverse direction to recover the secret message $M'$. Here, we use dense block in our $\phi(\cdot)$, $\rho(\cdot)$ and $\eta(\cdot)$ modules. The forward process and the reverse process of INN share the same network parameters and modules.}
 \label{fig:network}
\end{figure*}

\section{Introduction}
\label{sec:intro}
Digital watermarking has been widely used in the copyright protection of multimedia products since its inception \cite{van1994digital}. Digital watermarking hides the watermark information with specific meanings in digital content, such as images, videos, audio, documents, etc., through digital embedding. The extraction and recovery of watermark information can be used to prove the ownership and as evidence for identifying illegal infringement.
The goal of digital watermarking is to embed the secret message into the cover image in an invisible way and to extract the accurate secret message in the case of various distortions. In other words, digital watermarking requires high robustness and high invisibility.
Least significant bits (LSB) \cite{van1994digital} was the earliest research on image information hiding, which encodes the secret message on the least significant bits of image pixels. However, statistical measures \cite{fridrich2001detecting,fridrich2002practical,dumitrescu2002detection} can easily detect the secret information hidden by LSB.  Furthermore, the researchers find that watermarking in the frequency domain is more robust than the spatial ones. However, these traditional methods are heavily dependent on shallow manual image features, which imply that they need to be carefully designed and can not fully use the redundant information of cover images, so the robustness of such methods is limited. 

In recent years, with the upsurge of deep learning, many researchers have applied deep neural networks (DNN) to digital image watermarking, which significantly facilitates its development. These DNN-based methods \cite{zhu2018hidden,liu2019novel,jia2021mbrs} have shown advantages in robustness under various distortions compared with traditional methods. Zhu \textit{et al.} \cite{zhu2018hidden} proposed the first DNN-based method named Hidden and demonstrated superior performance than most traditional methods. Meanwhile, various subsequent DNN-based methods have adopted a similar framework. Such a framework uses a separate encoder and decoder, which treats the watermark encoding and decoding processes independently. Xu \textit{et al.} \cite{xu2021compact} simply applied invertible neural networks (INN) in image watermarking, which did not achieve satisfactory performance. Since higher robustness is the eternal pursuit of these DNN-based image watermarking approaches, our research question is: \textit{how to deep boost the robustness of DNN-based image watermarking under various distortions?}

In this paper, we propose a novel end-to-end digital image watermarking framework \tool to deep boost the robustness of DNN-based image watermarking.
The key novelty is the synergy of invertible neural networks (INN) and effective watermark features generation. The framework generates watermark features with redundancy and error correction ability through the effective neural network based message processor, synergized with the powerful information embedding and extraction abilities of INN to achieve higher robustness and invisibility. The powerful learning ability of neural networks enables the message processor to adapt to various distortions. We embed and extract watermark features by INN, in which the forward and inverse processes of INN share the same parameters and modules. Instead of hiding information directly in the spatial domain, we embed watermark information in the frequency domain of the cover image through the DWT. Moreover, we embed most of the watermark information into the high-frequency component through the $LL$ sub-band loss to improve invisibility from the perspective of the human visual system. 
 Experiment results show that our \tool framework outperforms existing SOTA (State Of The Art) methods on robustness and invisibility evidently.

In summary, the main contributions of this paper are:
\begin{itemize}
\item  We present a novel end-to-end digital image watermarking framework \tool to deep boost the robustness of DNN-based image watermarking, which is the synergy of the invertible neural networks (INN) and effective watermark features generation. 
\item We propose the neural network based message processor to generate watermark features with redundancy and error correction ability, significantly improving robustness against various distortions simultaneously. 
\item We propose  low-low ($LL$) sub-band loss to embed more watermark information in the high-frequency component of the discrete wavelet transform (DWT) domain, which enhances the invisibility of our \tool evidently.
\item We conduct extensive experiments to demonstrate that our framework \tool achieves higher robustness and invisibility than state-of-the-art approaches under various distortions.
\end{itemize}

The rest of the paper is organized as follows. We review the related work about digital watermarking  and INN in Section \ref{sec:related}. The details of the proposed framework are described in Section \ref{sec:methods}. Extensive experiments are presented in Section \ref{sec:experiments}. Finally, we conclude our work in Section \ref{sec:conclusin}.

\section{Related Work}

\label{sec:related}
\subsection{Digital Watermarking}
As a primary technology for copyright protection of content, digital watermarking is a popular research area in a wide range of real-world scenarios \cite{bender1996techniques,cox2002digital,hamidi2015blind,kumaraswamy2020digital,meng2018design,khadam2019digital}. The prior research of traditional methods has mainly investigated pixel-level manipulation \cite{van1994digital,bamatraf2010digital} in spatial domain. In order to improve the robustness, the traditional methods targeted the frequency domain, such as discrete Fourier transform (DFT) \cite{ruanaidh1996phase} domain, discrete cosine transform (DCT) \cite{hamidi2018hybrid} domain and discrete wavelet transform (DWT) \cite{guo2003digital} domain. 

In recent years, DNN-based methods have shown more advantages than traditional methods in both invisibility and robustness against various distortions, which is due to the powerful feature extraction ability of deep neural networks.
Zhu \textit{et al.} \cite{zhu2018hidden} first proposed a DNN-based framework to jointly trained the encoder and the decoder with a noise layer. Ahmadi \textit{et al.} \cite{ahmadi2020redmark} introduced domain transform in the DNN-based method and used the strength factor to change the strength of the watermark presented in the image. Luo \textit{et al.} \cite{luo2020distortion} proposed an attack network to simulate real distortions without any prior knowledge on the type of distortion during training, which obtained comparable results than other DNN-based methods and showed good generalization to unknown distortions. To achieve high robustness to the anticipated distortions, adding noise to encoded images during training is the most common and effective approach. However, this approach can not work well for non-differentiable distortions like JPEG compression. To overcome such limitation, many methods have been proposed. Zhu \textit{et al.} \cite{zhu2018hidden} and Shin \textit{et al.} \cite{shin2017jpeg} proposed differentiable JPEG-Mask and JPEG-SS respectively to approximate the real JPEG. However, the gap between the JPEG simulation and the real JPEG results in poor robustness against real JPEG. Liu \textit{et al.} \cite{liu2019novel} proposed a two-stage separable framework to solve the non-differentiable distortions problem. The method jointly trains the encoder and the decoder without a noise layer in stage one to obtain a powerful encoder. In stage two, the encoder is frozen and the decoder gains robustness enhancement from noise layer. Jia \textit{et al.} \cite{jia2021mbrs} focused on JPEG compression and proposed a mixed training method that randomly selects one from Identity, JPEG-Mask and real JPEG as noise layer during training. Furthermore, Zhang \textit{et al.} \cite{zhang2021towards} proposed the forward attack simulation layer method to improve the robustness against non-differentiable distortions in an end-to-end framework. The previous work has demonstrated that DNN have great potential in digital image watermarking.



\subsection{Invertible Neural Networks}
INN can be seen as a bijective function, which contains a forward mapping $f_{\theta}(\cdot)$ and a corresponding inverse mapping $f_{\theta}^{-1}(\cdot)$. Both mappings have a tractable Jacobian matrix, allowing explicit computation of posterior probabilities. Therefore, for the computational result that y generated by forward process $y = f_{\theta}(x)$, we can recover x directly through $x = f^{-1}_{\theta}(y)$.

Since the Invertible Neural Network was first proposed by Nice \cite{dinh2014nice}, INN has been applied for many computer vision tasks due to its outstanding performance. Zhu \textit{et al.} \cite{zhu2017unpaired} first utilized the bidirectional mapping of the INN to replace the cycle loss in CycleGan \cite{almahairi2018augmented}. Ardizzone \textit{et al.} \cite{ardizzone2019guided} proposed conditional INN for image generation and colorization. Xiao \textit{et al.} \cite{xiao2020invertible} introduced the invertible bijective transform of INN for image-rescaling tasks. Jing  \textit{et al.} \cite{jing2021hinet} and Lu \textit{et al.} \cite{lu2021large} applied INN to image steganography, which achieved large capacity and high invisibility.
In addition, INN are also used for image-to-video synthesis \cite{dorkenwald2021stochastic}, image compression \cite{wang2020modeling}, image denoising \cite{liu2021invertible} and video super-resolution \cite{zhu2019residual}. Although INN have great potential in information embedding and extraction, it shows weakness in robustness against lossy data compression and other distortions, which are the key issues of digital watermarking. Xu \textit{et al.} \cite{xu2021compact} applied INN in image watermarking, but the simple usage of INN can not achieve high robustness.


\section{The Framework \tool}
\label{sec:methods}
In this section, we describe the proposed framework \tool in detail. As shown in  Figure \ref{fig:network}, the proposed model is an end-to-end framework including DWT modules, message processors, INN blocks, and a noise layer. The following subsections will describe these modules in detail. Table \ref{tab:notation} presents the definition of notations in this paper.

\begin{table}[ht]
\caption{Notation definition. }
 \centering
\resizebox{\columnwidth}{!}{
 \begin{tabular}{@{}c|l@{}}
   \hline \hline
   Notation &   Definition \\
   \hline
  $M$ & secret message: the message to be hidden \\
  $M'$ & recovered message: the message recovered from encoded image\\
  $I_{co}$ & cover image: the image to hide secret message \\
  $I_{en}$ & encoded image: the image with secret message inside \\
  $I_{no}$ & noised image: the encoded image with noise \\
  $F_{co}$ & cover image features: the features of cover image \\
  $F_{en}$ & encoded image features: the features of encoded image\\
   $F_{no}$ & noised image features: the features of noised image\\
  $F_{M}$ & watermark features: the features of secret message\\
  $r$ & lost information: the information lost in encoding process\\
  $z$ & auxiliary matrix: the matrix to help recover secret message\\

   \hline \hline
 \end{tabular}
 }
 
 \label{tab:notation}
\end{table}

\subsection{Network Architecture}
Figure \ref{fig:network} shows the framework architecture of our \tool. In the watermark encoding process, the cover image and secret message are received as pair-wise input. The cover image $I_{co}$ $\in$ $\mathbb{R}^{C\times H \times W}$ is first transformed into frequency domain features $F_{co}$ $\in$ $\mathbb{R}^{4C\times H/2 \times W/2}$ through DWT), in which $C$ is channel number, $H$ is height, and $W$ is width. The secret message $M$ $\in$ $\left \{ 0,1 \right \} ^{L}$ needs to enter the message processor to generate the watermark features $F_{M}$ $\in$ $\mathbb{R}^{C_{M}\times H/2 \times W/2}$ with the same dimension as the frequency domain features $F_{co}$. $C_{M}$ is the channel number of watermark features, which is set to equal $L$ by default. Then, these two kinds of features go through a series of invertible neural network blocks. The output of the last block contains the frequency domain feature of encoded image $F_{en}$ and the lost information $r$. The encoded image features $F_{en}$ produce the encoded image $I_{en}$ after inverse discrete wavelet transform (IWT). The noise layer adds random noise to the encoded image and outputs the noised image $I_{no}$. In the watermark decoding process, the noised image features $F_{no}$ and an auxiliary matrix $z$ pass through a series of invertible neural network blocks to extract the watermark features $F_{M'}$, which are then fed into the inverse message processor to recover the secret message.

 \textbf{Discrete Wavelet Transform.}
DWT is a transform that decomposes the original image into different sub-bands, namely low-low ($LL$), low-high ($LH$), high-low ($HL$), and high-high ($HH$). The $LL$ sub-band contains no edge information since it is the low-frequency component of the image. According to the masking effect of the human sensory system, the digital watermark information can be embedded in the area where the original carrier is not easily perceptible, especially the high-frequency component in the DWT domain, so that the digital watermark has strong invisibility. Here, we use the $Haar$ wavelet to implement our DWT algorithm, which is simple and effective. After DWT, the feature map of the cover images with size $(C, H, W)$ is transformed into a frequency domain with size $(4C, H/2, W/2)$.

 \textbf{Message Processor.}
In the watermark encoding process, the one-dimensional secret message $M$ needs to be processed to two-dimensional watermark features $F_M$ with the same spatial dimension as the frequency domain features of cover images $F_{co}$. Here, we design a neural network to generate watermark features instead of simply duplicating the message or using a manual coding algorithm. $F_M$ contains much more bits than the original secret message, making it redundant and error-correcting. Here, we design a message processor to generate watermark features. First, the secret message $M$ $\in$ $\{0,1\}^L$ passes through a fully connected layer to increase the length of the message to $L'$. The motivation for this fully connected layer is from \cite{jia2021mbrs}, which uses the full connection layer to diffuse watermark information to the whole image to resist distortions. Here, we utilize the fully connected layer to diffuse the watermark features to the frequency domain of the cover image. Then, the message is reshaped to a $3$-dimension tensor $ \{0,1\}^{1\times h \times w}$ where $L' = h \times w$. It is fed into a single convolution layer to increase the channel number to $C_{M}$.  And then, the shape of $F_{M}$ expanded to $C_{M} \times H/2 \times W/2$ by several 2$\times$ upsampling layers. After that, the watermark features are sent to several Squeeze-and-Excitation Network (SENet) \cite{hu2018squeeze} blocks and then pass through INN concatenated with the cover image features.

 \begin{figure*}[ht]
 \centering
  \includegraphics[width=\linewidth]{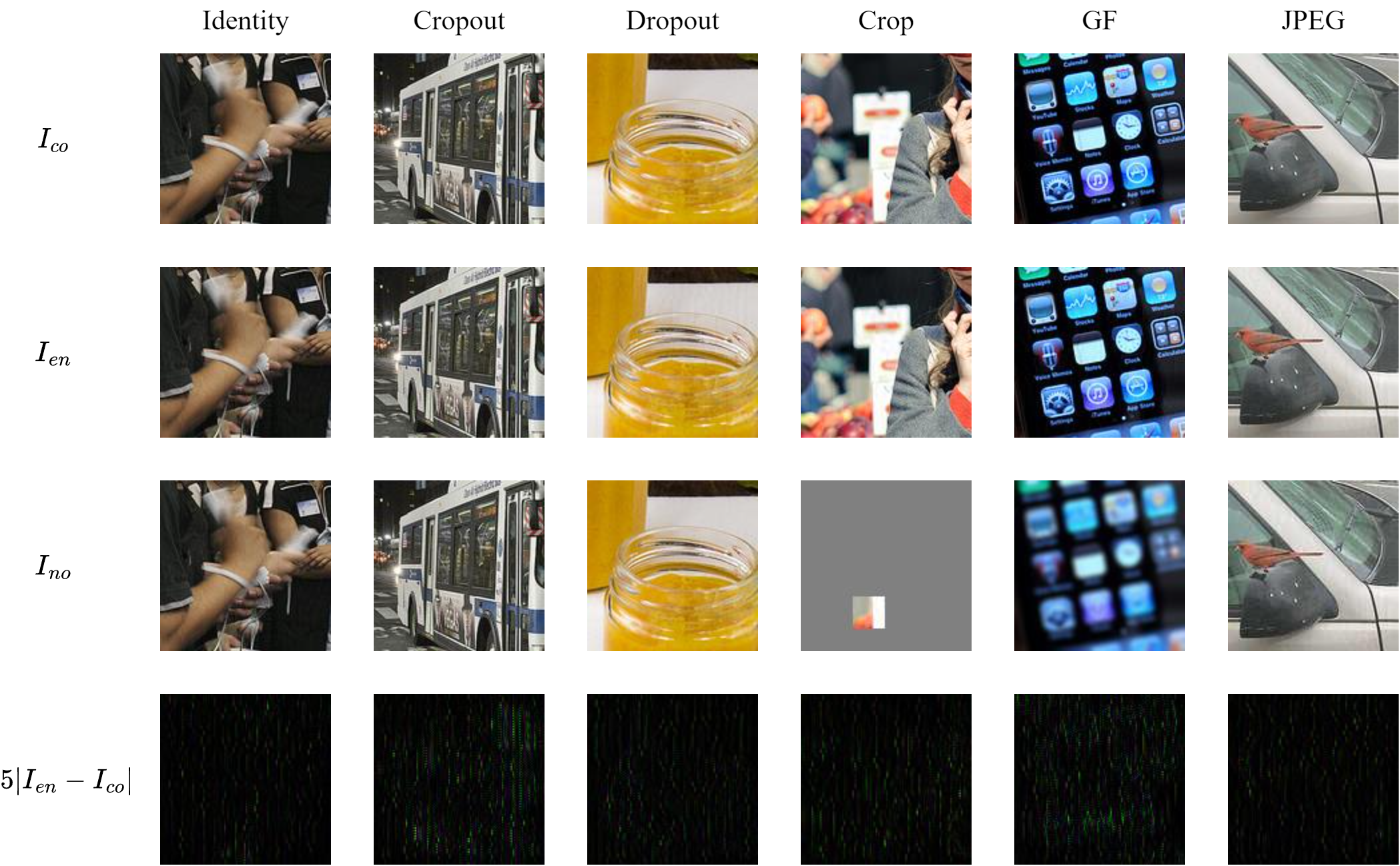}
  \caption{ Visual comparisons of cover and encoded images of our \tool. \textbf{First row}: Cover image $I_{co}$. \textbf{Second row}: Encoded images $I_{en}$. \textbf{Third row}: Corresponding noised images $I_{no}$ under six types of distortions . \textbf{Fourth row}: The difference between $I_{co}$ and $I_{en}$, which is five times magnified for visualization. PSNR = 39.98  and SSIM = 0.957. }
 \label{fig:result}
\end{figure*}

In the watermark decoding process, the inverse message processor receives the watermark features $F_{M'}$ from the first block of INN and outputs the recovered message $M'$. The inverse message processor has a similar network structure to the message processor but with a reverse information flow direction. During the upsampling and downsampling step, the number of upsampling or downsampling layers $n$ is decided by $L'$, $H$, and $W$. The formula is defined as follows,

\begin{equation}
     L' = h \times w  = (H/{2^{n-1}}) \times (W/{2^{n-1}}).
\end{equation}

 \textbf{Forward and Inverse Process in INN.}
The INN has shown outstanding performance in image steganography \cite{jing2021hinet,lu2021large}. In our proposed framework,  we build the INN blocks to embed and extract the watermark features to the image. The INN consists of $N$ INN blocks with the same structure.
As shown in Figure \ref{fig:network}, the forward and inverse processes in the INN share the same INN blocks but with reverse data flow directions.
In the forward process, the INN receives the frequency domain features of the cover image $F_{co}$ and watermark features $F_{M}$ as inputs.
For the $i$-th INN block, we denote the inputs as $b_1^i$ and $b_2^i$. The corresponding outputs $b_1^{i+1}$ and $b_2^{i+1}$ are calculated as follows,
\begin{equation}
b_1^{i+1}=b_1^i+\phi\left(b_2^i\right),
\end{equation}
\begin{equation}
   b_2^{i+1}=b_2^i \odot \exp \left(\alpha\left(\rho\left(b_1^{i+1}\right)\right)\right)+\eta\left(b_1^{i+1}\right), 
\end{equation}

where $\alpha(\cdot)$ is a $sigmoid$ function multiplied by a constant, $exp(\cdot)$ is an exponential function, and $\odot$ is the Hadamard product. Here, $\phi(\cdot)$, $\rho(\cdot)$ and $\eta(\cdot)$ are arbitrary functions and we apply the dense block for its simplicity and effectiveness. Finally, the INN outputs $b_1^{N+1}$ and $b_2^{N+1}$, where $b_1^{N+1}$ are the frequency domain features of encoded image $F_{en}$ and $b_2^{N+1}$ is lost information $r$. $F_{en}$ is then processed by IWT to generate the encoded image $I_{en}$.

Accordingly, the inverse process in the $i$-th block of INN is formulated as follows,

\begin{equation}
    b_1^{i}=b_1^{i+1}-\phi\left(b_2^i\right),
\end{equation}
\begin{equation}
b_2^i=\left(b_2^i-\eta\left(b_1^{i+1}\right)\right) \odot \exp \left(-\alpha\left(\rho\left(b_1^{i+1}\right)\right)\right).
\end{equation}
In order to keep the dimensions of the input and output of the INN consistent, the input contains an auxiliary matrix $z$ in addition to the frequency domain features $F_{no}$ generated by the noised image passing through the DWT module. The auxiliary matrix $z$ is randomly sampled from a Gaussian distribution.
Specifically, for the $N$-th block, the inputs $b_1^{N+1}$ and $b_2^{N+1}$ are the frequency domain features of the noised image $F_{no}$ and the auxiliary matrix $z$ correspondingly. Finally, the output of the first block $b_2^{1}$ is fed into the inverse message processor as the watermark feature $F_{M'}$.

 \textbf{Noise Layer.}
To improve the robustness against various distortions, inserting a noise layer after encoded images is the most common method in deep learning. However, the standard noise layer can not gain enhanced robustness from non-differentiable noise like JPEG. Zhang \textit{et al.} \cite{zhang2021towards} proposed a forward attack simulation layer method to overcome such limitations. The method calculates the difference between noised and encoded images and indicates this difference as the pseudo-noise. Then, the pseudo-noise is added to the encoded image to generate the pseudo-noised image. During the backward propagation, the pseudo-noise does not participate in the gradient propagation. Therefore, the gradient from the inverse blocks is directly back-propagated to the forward process in INN without passing through the noise layer. In our \tool, we propose to apply the forward attack simulation layer method for non-differentiable noise and standard noise layer for differentiable noise.

\subsection{Loss Functions}
Our \tool has two main goals. The goal of the watermark encoding process is to embed the secret message $M$ into the cover image $I_{co}$ to generate an encoded image $I_{en}$. The encoded image is required to be as similar as possible to the cover image. Therefore, the encoding loss function $\mathcal{L}_{en}$ is calculated by mean square error (MSE),
\begin{equation}
  \mathcal{L}_{en} = MSE(I_{co}, I_{en} ).  
\end{equation}

The watermark decoding process aims to extract and recover the secret message from the encoded image and minimize the difference between the original secret message $M$ and the recovered message $M'$. According to this goal, we define the decoding loss function as follows,
\begin{equation}
    \mathcal{L}_{de} = MSE(M, M').
\end{equation}

Since the human eye is less sensitive to noise in high-frequency sub-bands of the image, Zebbiche \textit{et al.} \cite{zebbiche2014efficient} proposed a perceptual masking model to embed the hidden information into the high-frequency components in the DWT domain, which enhances the performance significantly in terms of invisibility and robustness. Inspired by this work, we propose a $LL$ sub-band loss $\mathcal{L}_{LL}$ to enhance our \tool's invisibility. Suppose that $\mathcal{F}(\cdot)_{LL}$ is a filter to extract the $LL$ sub-band of the image. Here, we minimize the difference between the cover image and encoded image in $LL$ sub-band to allow more watermark information to be embedded in the high-frequency components of the encoded image. Therefore, the $LL$ sub-band loss is defined as follows,
\begin{equation}
    \mathcal{L}_{LL} = MSE(\mathcal{F}(I_{co})_{LL}, \mathcal{F}(I_{en})_{LL} ).
\end{equation}

In summary, the total loss function $\mathcal{L}_{total}$ is formulated as:
\begin{equation}
  \mathcal{L}_{total} = \mathcal{\lambda}_{en}\mathcal{L}_{en} + \mathcal{\lambda}_{de}\mathcal{L}_{de}+\mathcal{\lambda}_{LL}\mathcal{L}_{LL},  
\end{equation}

where $\mathcal{\lambda}_{en}$, $\mathcal{\lambda}_{de}$ and $\mathcal{\lambda}_{LL}$ are the weights of the corresponding losses for balancing different loss functions.

\begin{table*}[ht]
 \caption{Comparison with \cite{zhu2018hidden}, \cite{liu2019novel}, \cite{xu2021compact} and \cite{jia2021mbrs} trained by specified noise layer. SSIM is not reported in \cite{zhu2018hidden}, \cite{liu2019novel} and \cite{xu2021compact}. PSNR is measured for RGB channels, except in \cite{zhu2018hidden}, they use Y channel of YUV channels. Our \tool achieves the highest PSNR/SSIM and the lowest BER.}
 \centering
 \begin{tabular}{@{} c |c |c |c |c |c @{}}
  \hline
   \hline
  Method & Hidden \cite{zhu2018hidden} & TSDL \cite{liu2019novel}& IWN\cite{xu2021compact}  & MBRS \cite{jia2021mbrs} & \tool \\
   \hline
  Message size & 30 & 30 & 30 & 64 & 64 \\
  Noise Layer & JPEG-Mask & JPEG & JPEG-SS & Mixed(MBRS) & Forward ASL \\
  PSNR & (for Y) 30.09 & 33.51 & 36.16 & 36.49 & 39.75 \\
  SSIM & - & - & - & 0.9137 & 0.9486 \\
  BER & 15\% & 22.3\% & 8.8\% & 0.0092\% & 0.0018\% \\
   \hline
   \hline
 \end{tabular}
 \label{tab:jpeg}
\end{table*}

\begin{table*}[ht]
\caption{Comparison with \cite{zhu2018hidden}, \cite{liu2019novel}, \cite{xu2021compact} and \cite{jia2021mbrs} trained by combined noise layer. Our \tool achieves the lowest BER under six types of distortions. }
 \centering
  \resizebox{2\columnwidth}{!}{
 \begin{tabular}{@{} c |c |c |c |c |c |c |c@{}}
  \hline
   \hline
Method & PSNR & Identity & Cropout ($p = 0.3$) & Dropout ($p = 0.3$) & Crop ($p = 0.035$) & GF ($\sigma$ = 2) & JPEG ($Q = 50$) \\
   \hline
  Hidden \cite{zhu2018hidden} & 33.5 & 0\% & 6\% & 7\% & 12\% & 4\% & 37\% \\
  TSDL \cite{liu2019novel} & 33.5 & 0\% & 2.7\% & 2.6\% & 11\% & 1.4\% & 23.8\% \\
  IWN \cite{xu2021compact} & 33.0 & 0.06\% & 5.29\% & 24.71\% & 16.69\% & 13.89\% & 23.13\% \\
  MBRS \cite{jia2021mbrs} & 33.5 & 0\% & 0.0027\% & 0.0087\% & 4.15\% & 0.011\% & 4.48\% \\
  \tool & 34.8 & 0\% & 0.002\% & 0.0013\% & 3.74\% & 0\% & 0\%\\
   \hline
   \hline
 \end{tabular}
 }
 \label{tab:combine}
\end{table*}

\subsection{Strength Factor}
We can get the difference between the encoded image and cover image $I_{diff} = I_{en} - I_{co}$, which is the watermark mask with the secret message $M$. Therefore, we can adjust the watermarking strength by a strength factor $S$ to balance robustness and invisibility:
\begin{equation}
\label{eq:sf}
    I_{en, S} = I_{co} + S*I_{diff}.
\end{equation}
 In section \ref{factor}, we will discuss the influence of different strength factors on the performance of our framework.

\section{Experiments}
\label{sec:experiments}
In this section, we conduct experiments to validate the efficacy and robustness of our proposed framework \tool. Performance against the baseline and parameter sensitivity on performance are considered for effectiveness verification. The experiments cover a wide range of complex image distortions in terms of robustness.

\subsection{Experimental Settings}
 \textbf{Implementation Details.}
Our \tool is trained under $10000$ images of COCO \cite{lin2014microsoft} training data set, and the neural network is evaluated through $5000$ images of COCO test data set to ensure the generalization of the model. Our model is implemented through PyTorch \cite{collobert2011torch7} and runs on  NVIDIA A100 GPUs $\times 2$. In order to compare with previous work, we adjust the image's resolution to $128 \times 128$. The secret message lengths are $30$ bits and $64$ bits for combined training and specified training, where each bit is generated by random sampling. Our \tool contains $16$ invertible neural network blocks. The parameters $\mathcal{\lambda}_{en}$, $\mathcal{\lambda}_{de}$ and $\mathcal{\lambda}_{LL}$ are set to $0.1$, $100.0$ and $0.1$, respectively. The mini-batch size is $16$, and the Adam optimizer is adopted with standard parameters and an initial learning rate of $1 \times 10^{-5.0}$. For specified training, we add one specified distortion in the noise layer during the training. For combined training, we choose one distortion from Identity, Cropout ($p = 0.3$), Dropout ($p = 0.3$), Crop ($p = 0.035$), Gaussian Filter (GF, $\sigma = 2$) and JPEG ($Q = 50$) randomly and add it to encoded image in noise layer. Since the input of the watermark decoding process is a fixed size tensor, we pad the cropped images with pixels expressed as $(127,127,127)$ in RGB channels. 

\begin{table*}[ht]
 \caption{BER, PSNR, SSIM value under several strength factors. BER is tested under different distortions. As the factor increases, PSNR and SSIM values decrease steadily, but BER value decreases first and then increases.}
 \centering
 \begin{tabular}{@{} c |c |c |c |c |c |c |c |c |c @{}}
  \hline
   \hline
 \multicolumn{2}{c|}{Strength factor} & 0.5 & 0.75 & 1.0 & 1.25 & 1.5 & 1.75 & 2.0 & 2.25  \\
   \hline
  \multirow{6}*{BER} & Identity & 0.00\% & 0.00\% & 0.00\% & 0.00\%  & 0.00\%  & 0.00\%   & 0.00067\% & 0.0087\%  \\
  ~ & Cropout & 0.79\% & 0.015\% & 0.00067\% & 0.00067\% & 0.0013\%  & 0.0013\%   & 0.0013\% & 0.0013\%  \\
  ~ & Dropout & 1.67\%& 0.051\% & 0.0047\% & 0.00\% & 0.00\%   & 0.00067\%  & 0.00\%  & 0.00\%  \\
    ~ & Crop & 25.6\% & 12.5\% & 5.68\% & 2.92\%  & 1.96\%  & 1.62\%  & 1.56\% & 1.56\%    \\
      ~ & GF & 4.11\% & 0.081\% & 0.00\% & 0.00\% & 0.00067\%  & 0.002\%  & 0.0027\%  & 0.004\%  \\
        ~ & JPEG & 8.43\% & 0.175\%  & 0.0034\%  & 0.00067\% & 0.00\%  & 0.00\% & 0.00\% & 0.0013\%     \\
     \hline
   \multicolumn{2}{c|}{PSNR} & 44.34 & 40.81 & 38.32 & 36.38  & 34.79 & 33.46 & 32.30 & 31.27\\
      \hline
   \multicolumn{2}{c|}{SSIM} & 0.9801 & 0.9584 & 0.9321 & 0.9033  & 0.8733 & 0.8430 & 0.8130 & 0.7838\\
   \hline
   \hline
 \end{tabular}
 \label{tab:factor}
\end{table*}



\begin{figure*}
 \centering
  \includegraphics[width=\linewidth]{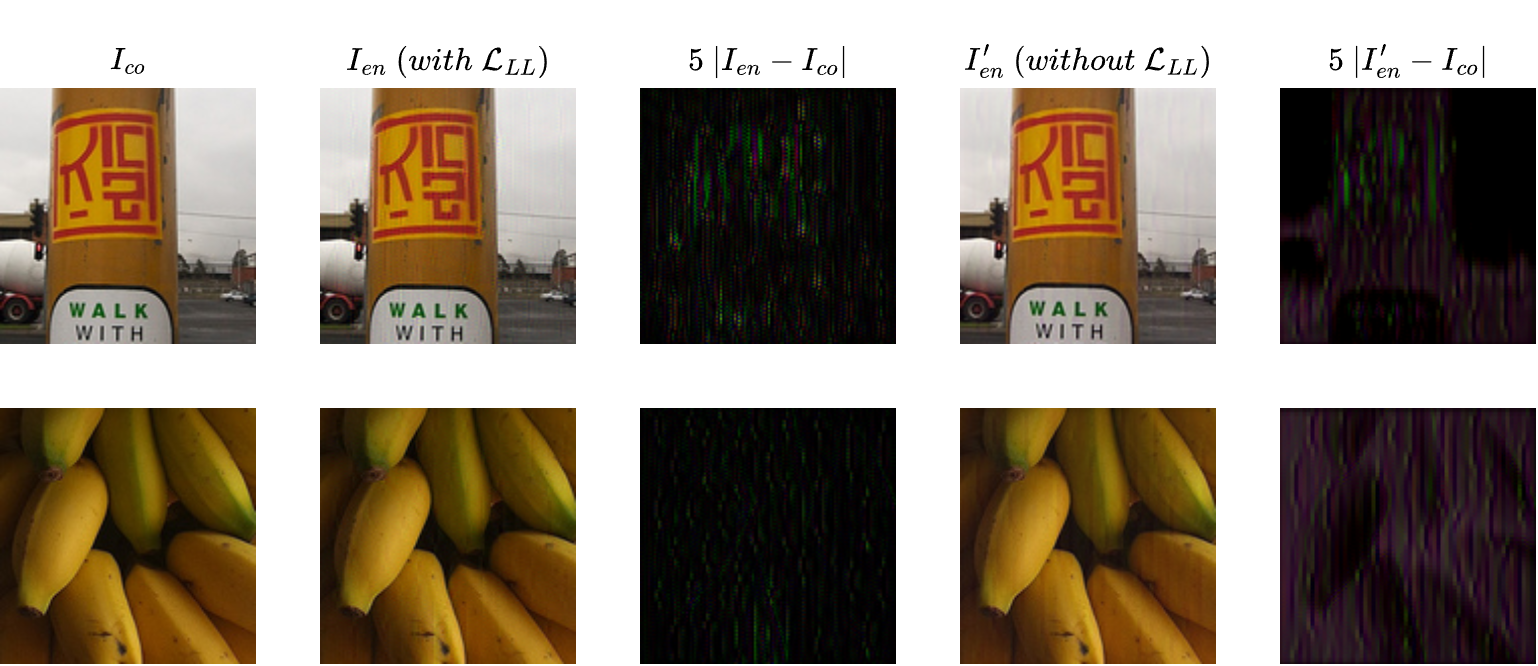}
  \caption{ Samples of encoded and cover images for our \tool with/without $\mathcal{L}_{LL}$ loss. From left to right are: the cover image $I_{co}$, the encoded image with $\mathcal{L}_{LL}$ loss $I_{en}$, the difference between $I_{en}$ and $I_{co}$, the encoded image without $\mathcal{L}_{LL}$ loss $I'_{en}$, the difference between $I'_{en}$ and $I_{co}$. The differences are five times magnified for visualization. }
 \label{fig:lose}
\end{figure*}

  \textbf{Metrics.}
There are two objectives in digital image watermarking: robustness, measured by the Bit Error Rate (BER) of the recovered message;
and invisibility, measured by both peak signal-to-noise ratio (PSNR) and structural similarity (SSIM) between the cover and encoded image.

  \textbf{Benchmarks.}
To verify the performance of our proposed \tool, we conduct the comparison with four DNN-based methods\cite{zhu2018hidden}, \cite{liu2019novel}, \cite{xu2021compact} and \cite{jia2021mbrs}, especially with the state-of-the-art (SOTA) digital watermark method proposed in \cite{jia2021mbrs}. The architecture of \cite{xu2021compact} is based on INN, but the author does not open-source their code. Although \cite{zhu2018hidden} and \cite{liu2019novel} open-sourced their code, we can not reproduce the results as they reported. In order to respect the results they have reported, we directly compare them with the results published in \cite{zhu2018hidden}, \cite{liu2019novel} and \cite{xu2021compact}. \cite{jia2021mbrs} open-sourced both their codes and their model, so we use the pre-trained model for comparison. 

\subsection{Experimental Results}
In this subsection, we show the comparison results between our \tool and previous methods. For JPEG compression, since the message length of each method is different, we use a larger message length to train our model for a fair comparison. Specifically, we use secret message length $L = 64$, which is larger than \cite{zhu2018hidden}, \cite{liu2019novel} and \cite{xu2021compact}. For combined compression, we use secret message length $L = 30$. In terms of visual quality, our watermark mask is distributed across the image like vertical stripes. The qualitative comparison between the cover image and the encoded images, and the illustration of the noised image under different distortions are shown in Figure \ref{fig:result}.

\subsubsection{JPEG Comparison}
In this experiment, we mainly compare the JPEG robustness between the proposed method and four other methods, \cite{zhu2018hidden}, \cite{liu2019novel}, \cite{xu2021compact} and \cite{jia2021mbrs}. Our model is trained by the forward ASL method proposed by \cite{zhang2021towards} with real JPEG. All the testing processes are performed under real JPEG compression with quality factor $Q=50$. As shown in Table \ref{tab:jpeg}, our \tool achieves higher PSNR and SSIM and lower BER, which indicates higher invisibility and robustness. Significantly, the BER is only $0.0018\%$. Experimental results demonstrate that our \tool is more effective against JPEG compression than the previous methods.

\subsubsection{Combined Comparison}
To demonstrate that our \tool is resistant to various distortions simultaneously, we train a noise-resistant combined model by using a random noise layer for each mini-batch during the training. The noise layers include Identity, Cropout ($p = 0.3$), Dropout ($p = 0.3$), Crop ($p = 0.035$), Gaussian Filter (GF, $\sigma = 2$) and JPEG ($Q = 50$). We use forward ASL for JPEG since it is a non-differentiable distortion. We adjust the strength factor to achieve the best performance of our \tool. When the strength factor is large, with the increment of the strength factor, the robustness and invisibility of the encoded image decrease simultaneously. Therefore, we do not use a large strength factor to obtain similar PSNR value with different methods. As we can see in Table \ref{tab:combine}, our \tool can maintain higher PSNR and lower BER than previous methods under each kind of distortion. In particular, our method can achieve $0\%$ bit error rate for Gaussian filter and JPEG compression distortions. Furthermore, our \tool has a dramatic improvement in robustness compared to \cite{xu2021compact}, which is also implemented based on INN.

\subsubsection{Strength Factor}
\label{factor}
We introduce a strength factor $S$ in Equation ~\eqref{eq:sf}  to adjust the trade-off between the quality of the encoded image and the BER of the recovered message. In order to determine the best strength factor for our method, we test our \tool under different strength factor values from $0.5$ to $2.25$. As we can see in Table \ref{tab:factor}, with the increment of strength factor, PSNR and SSIM values decrease steadily. However, the BER values under different distortions decrease first and then increase, which is different from the negative correlation relationship between $S$ and BER under different quality factors for JPEG compression in \cite{jia2021mbrs}. This observation indicates that the message recovery in our \tool depends not only on those changed pixels but also on those unchanged pixels. Therefore, it demonstrates that our \tool can better utilize the redundant information of the image. 

\subsubsection{Effectiveness of $LL$ sub-band loss}
The $\mathcal{L}_{LL}$ loss is designed to hide more watermark information in the high-frequency component of the cover image so that the human eye can hardly detect the presence of a watermark.
Visual comparisons of encoded images with and without $\mathcal{L}_{LL}$ loss are shown in Figure \ref{fig:lose}. As we can see, the encoded images with $\mathcal{L}_{LL}$ loss maintain a higher image quality rather than the one without $\mathcal{L}_{LL}$ loss, which demonstrates that the $\mathcal{L}_{LL}$ loss can enhance the invisibility of our method. In addition, the differences between $I_{co}$ and $I'_{en}$ are like the afterimage of the cover image. In contrast, we can only see some sparse vertical artifacts in the differences between $I_{co}$ and $I_{en}$.



\section{Conclusion}
\label{sec:conclusin}
In this paper, we have proposed a novel end-to-end digital image watermarking framework \tool to deep boost the robustness of DNN-based image watermarking. 
We have proposed to generate robust watermark features through message processing against distortions. In model training, the $LL$ sub-band loss has been proposed to improve the visual quality of the encoded image.
Our \tool has synergized the strengths of INN, discrete wavelet transform and the forward attack simulation layer method to offer better robustness and invisibility. 
Extensive experiments have demonstrated that our \tool can achieve digital image watermarking with higher robustness and invisibility, outperforming other SOTA methods evidently.

{\small
\bibliographystyle{ieee_fullname}
\bibliography{egbib}
}

\end{document}